# FUSegNet: A Deep Convolutional Neural Network for Foot Ulcer Segmentation

**Mrinal Kanti Dhar\*, Taiyu Zhang, Yash Patel, Sandeep Gopalakrishnan, and Zeyun Yu**

Big Data Analytics and Visualization Laboratory, Department of Computer Science
University of Wisconsin-Milwaukee, Milwaukee, WI 53201, USA

mdhar@uwm.edu, taiyu@uwm.edu, yspatel@uwm.edu, sandeep@uwm.edu, and yuz@uwm.edu

\*Corresponding author

**ABSTRACT**

This paper presents FUSegNet, a new model for foot ulcer segmentation in diabetes patients, which uses the pre-trained EfficientNet-b7 as a backbone to address the issue of limited training samples. A modified spatial and channel squeeze-and-excitation (scSE) module called parallel scSE or P-scSE is proposed that combines additive and max-out scSE. A new arrangement is introduced for the module by fusing it in the middle of each decoder stage. As the top decoder stage carries a limited number of feature maps, max-out scSE is bypassed there to form a shorted P-scSE. A set of augmentations, comprising geometric, morphological, and intensity-based augmentations, is applied before feeding the data into the network. The proposed model is first evaluated on a publicly available chronic wound dataset where it achieves a data-based dice score of 92.70%, which is the highest score among the reported approaches. The model outperforms other scSE-based UNet models in terms of Pratt's figure of merits (PFOM) scores in most categories, which evaluates the accuracy of edge localization. The model is then tested in the MICCAI 2021 FUSeg challenge, where a variation of FUSegNet called x-FUSegNet is submitted. The x-FUSegNet model, which takes the average of outputs obtained by FUSegNet using 5-fold cross-validation, achieves a dice score of 89.23%, placing it at the top of the FUSeg Challenge leaderboard. The source code for the model is available on GitHub.

## 1 Introduction

Chronic wounds are those that fail to progress through the normal healing process or for which the healing process does not restore anatomic and functional integrity after three months [1]. Among different lower extremity chronic wounds, diabetic foot ulcer (DFU) is very prevalent. Long-term diabetic patients may develop neuropathy, a result of nerve damage brought on by chronically raised blood glucose levels. Neuropathy with or without peripheral vascular disease reduces or completely diminishes the ability to feel pain in the feet, leading to an ulceration called diabetic foot ulcer (DFU) which can range in depth from superficial to deep.

Obese and diabetic patients are more vulnerable to the development of chronic wounds. 34.2 million Americans and 463 million other people worldwide have diabetes which is expected to increase by 25% in 2030 [2]. The risk of developing a foot ulcer in a diabetes patient is 30% with up to 85% chance that the foot ulcer will precede lower limb amputation in diabetes [3]. Foot ulcers have both social and economic impacts. Such chronic wounds can impact the patient's quality of life. They can cause serious consequences, including limb amputations and death if they are not treated appropriately [4]. According to research using the Medicare 5% Limited Data Set for CY2014, nearly 15% of Medicare beneficiaries (8.2 million) are affected with chronic wounds, which have an annual cost to Medicare of between $28.1 and $31.7 billion, with diabetic wound infections being the prominent prevalence category (3.4%) apart from the surgical infections [5].

To evaluate and manage chronic wounds, track the wound healing process, and plan for future interventions, the wound area must be precisely measured [6]. Manual measurement of the wound region suffers from three limitations

– (1) costly in terms of time and labor, (2) needs medical experts and (3) error-prone. Additionally, the coronavirus disease (COVID) pandemic in 2020 severely affected global health care, including wound care [7]. An alternative is to apply computer-aided methods to segment the wound regions. Computerized methods offer the following advantages – (1) faster and more efficient, (2) automatic, (3) easier to extract additional morphological features (such as height, width, depth, area, etc.), and (4) easier to keep digital records.

## 2    Literature Review

Literature available for diabetic foot ulcer (DFU) segmentation can be divided into two categories. The first category deploys traditional image processing techniques and machine learning. The second category uses various deep learning methods. Song et al. [8] used four segmentation techniques, k-means, edge detection, thresholding, and region growing, to extract features from DFU images. They optimized parameters using Grid search and the Nelder-Mead simplex algorithm. They then used a Multi-Layer Perceptron (MLP) and a Radial Basis Function (RBF) neural network to identify the wound region. Wantanajittikul et al. [9] applied Cr-Transformation and Luv-Transformation to highlight the wound region removing the background. A pixel-wise Fuzzy C-mean Clustering (FCM) technique is used to segment the transformed images. Jawahar et al. [10] compare three methods for DFU segmentation – mask-based segmentation, L*a*b* color space-based segmentation, and K-means clustering-based segmentation. They experimented on the Medetec dataset [11] consisting of 152 clinical images and got the best segmentation result for K-means clustering. Heras-Tang et al. [12] used a logistic regressor model to classify ulcer region pixels followed by a post-processing stage that applies a DBSCAN clustering algorithm, together with dilation and closing morphological operators. They achieved an F1-score of 0.88 on a dataset having 26 images for training and 11 for validation.

However, the above-mentioned methods suffer from at least one of the following limitations: (1) require some extent of feature engineering, (2) sensitive to skin color, illumination, and resolution, (3) require manual tuning of parameters, (4) not fully automatic end-to-end, (5) not evaluated on a large dataset. These limitations can be fully or partially overcome by deploying deep learning models.

Goyal et al. [13] introduced a foot ulcer dataset consisting of 705 and achieved a dice coefficient of 79.4% using FCN-16 architecture. The network tends to create smooth contours; therefore, its segmentation accuracy is restricted in identifying small wounds and wounds with uneven borders. Liu et al. [14] proposed a framework called WoundSeg that uses MobileNet architecture with different numbers of channels alongside VGG16 architecture. On their dataset of 950 photos captured in an uncontrolled lighting setting with a complicated background, they achieved a Dice accuracy of 91.6%. However, rather than using experts, a watershed method is used to semi-automatically annotate their dataset. Wang et al. [15] used lightweight MobileNetv2 on a chronic wound dataset consisting of 810 training images and 200 test images. They added a post-processing step to fill the gaps left by the presence of abnormal tissue and remove small regions. They achieved a data-based Dice score of 90.47%. Cao et al. [16] classified wound images into five grades using the Wagner diabetic foot grading method and used mask R-CNN for semantic segmentation. They had a dataset of 1426 DFU images, with 967 images having nested labels and 459 images having single-graded labels. Their model had an accuracy of 98.42% but did not significantly improve the F1-score compared to region proposal-based methods. Additionally, their model is sensitive to feature vector concatenation, without which its performance decreases. Ramachandram et al. [17] developed an attention-embedded encoder-decoder network for wound tissue segmentation. The model consisted of two stages: the first stage segmented the wound region, and the second stage segmented the four wound tissue types (epithelial, granulation, slough, and eschar). The model was trained on 467,000 images for wound segmentation and 17,000 images for tissue segmentation, using the largest wound dataset reported so far. They evaluated the model on a dataset of 58 images and found poor performance for epithelial tissue. Huang et al. [18] first detected the wound region by Fast R-CNN and then applied GrabCut and SURF algorithms to determine the wound boundaries. Consequently, the segmentation part is based on classical image-processing techniques rather than deep learning. The accuracy was 89%, though the mean average precision (mAP) was limited to 58. Additionally, the GrabCut algorithm incorporates GMM data and computes iterative minimization, which includes some random information and produces marked contours that are less precise in practice. Mahbod et al. [19] proposed an ensembled network for the FUSeg challenge 2021 [20] consisting of LinkNet and U-Net using pretrained EfficientNetB1 and EfficientNetB2 encoders, respectively, with additional pretraining using the Medetec dataset [11]. The FUSeg dataset consists of 1210 DFU images where 1010 images are provided for training and 200 images for evaluation. They achieved data-based Dice scores of 88.80% and 92.07% for the FUSeg dataset and the chronic wound dataset, another dataset of the same organizer, respectively. However, the segmentation performance deteriorated when there was no wound or a very small wound region. Kendrick et al. [21] used a dataset called DFUC2022 for diabetic foot ulcer (DFU) segmentation, consisting of 2000 training and 2000 testing images, which is reported as the largest DFU dataset available currently. They proposed a network using FCN32 with a modified VGG backbone that replaced ReLU activation with Leaky-ReLU and removed the bottom three max-pooling layers to prevent excessive downsampling. To address the class imbalance, they trained on patches that had DFU regions and used a patch size of 64x48 with a stride of 32x24. They achieved a dice score of 74.47%. Yi et al. [22] and Hassib et al. [23] utilized the same DFUC2022 dataset for diabetic foot ulcer segmentation. In their work, Yi et al. [22] proposed a novel method employing OCRNet with a ConvNeXt backbone. Additionally, they introduced a boundary loss function that calculated binary cross-entropy loss between the boundary maps generated from the ground truth mask and predictions. This approach resulted in a Dice score of 72.80%. In the study by Hassib et al. [23], SegFormer MiT-B5 was applied to the DFUC2022 dataset, achieving a Dice score of 69.89%. Attempts were made to enhance segmentation performance through ensemble methods involving SegFormer and DeepLabV3+, but no improvement was observed. Lien et al. [24] focused on segmenting only the granulation tissue rather than the entire wound region. They conducted experiments on 219 images from 100 patients, dividing each image into 32×32 patches. ResNet18 was then applied to classify each patch into three categories: granulation, non-granulation, and non-wound patch, achieving an Intersection over Union (IoU) score of 60%. Kairys et al. [25] conducted a review of publications on artificial intelligence-based ulcer detection and segmentation from 2018 to 2022. Article selection was performed using the Preferred Reporting Items for Systematic Reviews and Meta-Analyses (PRISMA) methodology.

In this study, our contribution can be summarized as follows –

1. We propose a modified squeeze-and-excitation (SE) attention module called parallel scSE (P-scSE) that combines both the additive and max-out spatial and channel squeeze-and-excitation (scSE) modules.
2. We propose a noble encoder-decoder-based architecture called FUSegNet for foot ulcer segmentation. The encoder path incorporates a pretrained EfficientNet-b7. In each decoder stage, we integrate P-scSE modules for fusion. Additionally, we develop a modified version called x-FUSegNet (pronounced cross-FUSegNet) that takes the average of outputs obtained by the FUSegNet using 5-fold cross-validation.
3. We propose a new arrangement for the proposed attention model. As opposed to [26], instead of using it at the end of each decoder stage, we use it in the middle followed by a 3×3 Conv-ReLU-BN. Layer-wise visualization shows that such an arrangement smooths the output obtained by the attention module.
4. We evaluate the proposed FUSegNet model on the chronic wound dataset [15] and the FUSeg Challenge 2021 dataset [20]. We first carry out extensive experiments on the chronic wound dataset to finalize all the network parameters. We then participate in the FUSeg Challenge 2021 and apply x-FUSegNet to evaluate the performance.
5. The FUSegNet model outperforms state-of-the-art methods for the chronic wound dataset with a dice score of 92.70%. The x-FUSegNet is currently at the top of the leaderboard of the FUSeg Challenge 2021 with a dice score of 89.23% [27].

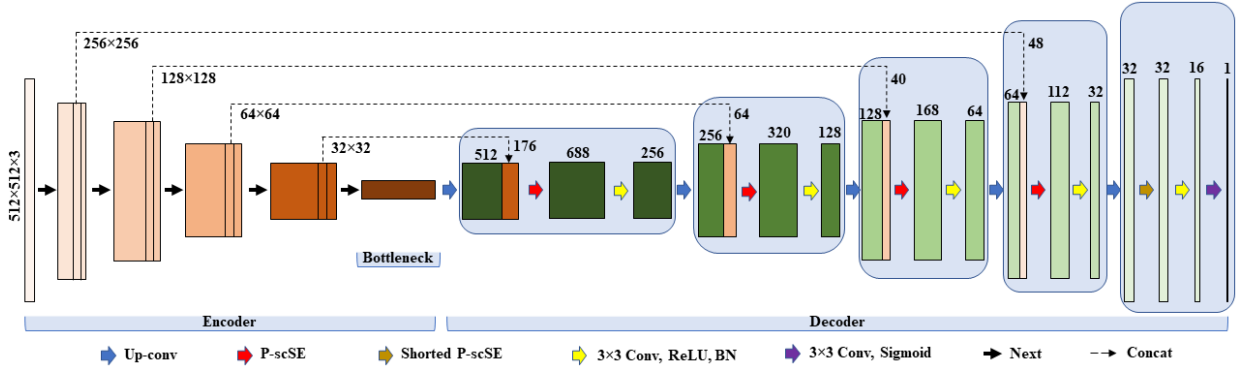

**Figure 1** Overview of the proposed FUSegNet architecture. Numbers are shown for an input of 512×512×3.

## 3  Materials and Methods

### 3.1  Datasets

In this paper, two datasets were used to evaluate our models. They are – the chronic wound dataset [15] and the FUSeg Challenge 2021 dataset [20].

The chronic wound dataset is a publicly available dataset containing 1010 images of foot ulcers with a resolution of 224 × 224. It was generated by the Big Data Analytics and Visualization Lab – UWM in collaboration with the AZH Wound and Vascular Center, Milwaukee, Wisconsin, USA. The dataset is divided into 810 images for training and 200 images for testing, all taken from 889 patients under uncontrolled lighting conditions using iPad Pro and Canon SX 620 HS digital cameras. A YOLOv3 object detection model [28] was used to locate the wound region, and the images were manually labeled to create binary segmentation masks verified by wound care experts.

The FUSeg dataset is an extension of the chronic wound dataset and was created by the same group. It contains 1210 foot ulcer images, of which 1010 are identical to the chronic wound dataset but now have the entire view instead of just the wound region. These 1010 images are used for training, and the remaining 200 images are for testing, all with a fixed size of 512 × 512. The FUSeg dataset is used for the MICCAI 2021 Foot Ulcer Segmentation Challenge, with the segmentation masks for the test images only used for evaluation and kept private by the organizers.

### 3.2  FUSegNet architecture

Extensive experimentation was conducted on the chronic wound dataset before taking part in the FUSeg Challenge 2021, which led to the development of a new architecture named FUSegNet. The overview of the proposed network architecture is illustrated in Figure 1. It is primarily an encoder-decoder-based architecture. It incorporates EfficientNet-b7 and a decoder that embeds our proposed modified attention module (P-scSE). In summary, the encoder is a way of down-sampling that collects semantic or contextual information. The decoder, on the other hand, is an up-sampling path that restores spatial information. The necessary high-resolution (but low semantic) information is finally sent from the encoder to the decoder through shortcut connections between two paths. A modified attention mechanism called parallel spatial and channel squeeze-and-excitation (P-scSE) is fused in the decoder path.

#### 3.2.1  Why EfficientNet-b7

To avoid manual scaling, an EfficientNet architecture is employed as the encoder backbone in this study. Convolutional neural networks rely heavily on scaling, which can be achieved in various ways, including depth-wise, widthwise, and resolution-wise scaling. However, traditional scaling methods are random and require manual tuning, making them time-consuming and challenging to perform simultaneously. The authors [29] of the EfficientNet propose a novel architecture that uniformly scales depth, width, and resolution using fixed scaling coefficients ($\alpha$, $\beta$, and $\gamma$) and a compound coefficient $\phi$. Mathematically, depth, width, and resolution scaling are achieved by $\alpha^\phi$, $\beta^\phi$, and $\gamma^\phi$, respectively. To determine the appropriate values of these coefficients, the authors used neural architecture search (NAS) to construct a baseline network called EfficientNet-B0, with values of $\alpha$, $\beta$, and $\gamma$ set at 1.2, 1.1, and 1.15, respectively. After a small grid search under the constraint of $\alpha \cdot \beta^2 \cdot \gamma^2 \approx 2$, the baseline model was scaled up with different values of $\phi$ to obtain EfficientNet-B1 to B7. EfficientNet-B0, the baseline model, has an image resolution of 224×224, while the value of $\phi$ in EfficientNet-B7 is 6. As a result, the resolution of EfficientNet-B7 can be calculated as $224 \times \gamma^\phi = 224 \times 1.15^6 \approx 518$. In this work, EfficientNet-B7 is chosen as its resolution suitable for the FUSeg Challenge 2021 which has an image resolution of 512×512. Additionally, EfficientNet-B7 demonstrated exceptional performance on ImageNet.

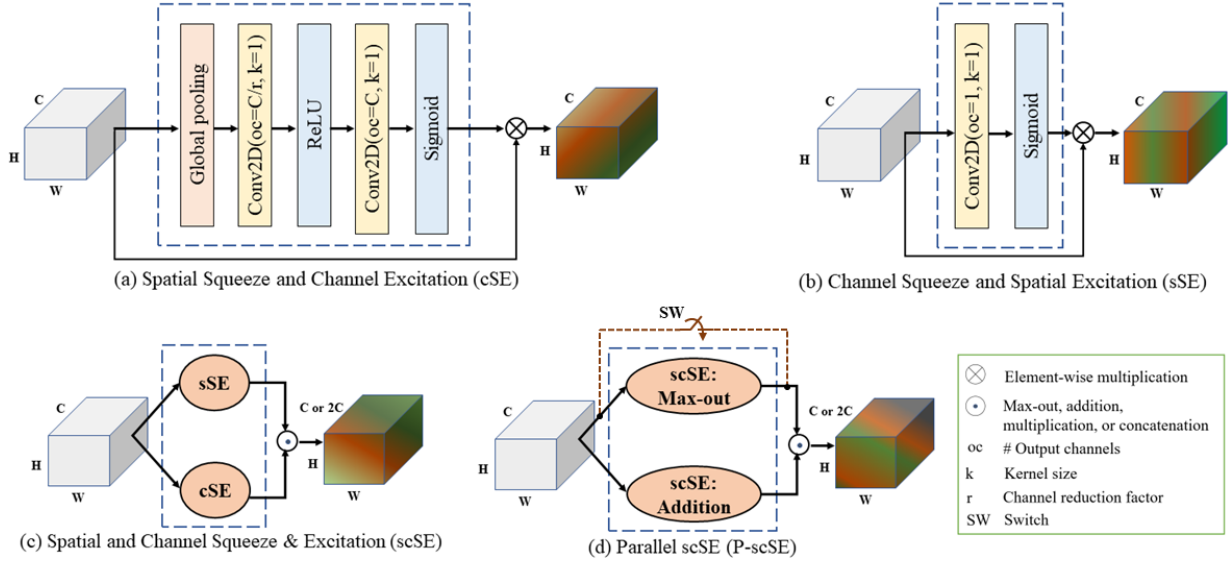

**Figure 2** Parallel scSE (P-scSE) module.

*3.2.2 Parallel spatial and channel squeeze-and-excitation (P-scSE)*

In this section, we discuss our proposed modified squeeze-and-excitation (SE) module called Parallel spatial and channel squeeze-and-excitation (P-scSE). Figure 2 illustrates the formation of the P-scSE module. The SE module was first proposed by Hu et al. [30] to enhance network representational power by emphasizing informative features while suppressing less useful ones. It generates a channel descriptor using global average pooling and excites channel-wise dependencies. It is also termed *cSE* as it excites along the channel axis (see Figure 2(a)). Roy et al. [26] introduced the sSE module (see Figure 2(b)), which squeezes along the channel axis while exciting spatially, and the scSE module (see Figure 2(c)), which combines the cSE and sSE blocks to aggregate spatial and channel-wise information. The notation used is 's' for spatial and 'c' for the channel, while 'S' and 'E' represent squeeze and excitation, respectively. These modules are useful for complex anatomy tasks like medical image segmentation.

Let $\mathbf{X}^{[l-1]}$ be the output feature map of the ($l$-1)-th level of the decoder. Then the intermediate feature map at the $l$-th level, $\mathbf{X}_i^{[l]} = \mathbf{X}_u^{[l]} \frown \mathbf{X}_s^{[l]}$, where, $\mathbf{X}_i^{[l]} \in \mathbb{R}^{H \times W \times C}$, will be achieved by doing concatenation ($\frown$) between $\mathbf{X}_u^{[l]}$, which is achieved by upsampling $\mathbf{X}^{[l-1]}$, and $\mathbf{X}_s^{[l]}$, which is the encoder output at the $l$-th level passed through a skip connection.

This intermediate $\mathbf{X}_i^{[l]} = [\mathbf{x}_{i,1}, \mathbf{x}_{i,2}, \dots, \mathbf{x}_{i,C}]$ is passed through the *P-scSE* module. Before diving into the P-scSE module, let's consider $\mathbf{X}_i^{[l]}$ is passed through the *scSE* module. During the channel excitation (cSE), the channel descriptor, $\mathbf{X}_{sS}^{[l]} \in \mathbb{R}^{1 \times 1 \times C}$ is achieved by using global average pooling, which, for channel $c$, can be expressed as:

$$x_{sS,c}^{[l]} = F_{sS}(x_{i,c}) = \frac{1}{H \times W} \sum_{p=1}^{H} \sum_{q=1}^{W} x_{i,c}^{[l]}(p,q) \tag{1}$$

Where $\mathbf{F}_{sS}(.)$ is the squeezing operator along the spatial dimensions. It is then passed through the channel excitation which is expressed as:

$$\mathbf{X}_{cE}^{[l]} = \mathbf{F}_{cE}(\mathbf{X}_{sS}^{[l]}, \mathbf{W}) = \sigma_2(\mathbf{W}_2 \mathbf{X}_{cE,i}^{[l]}) = \sigma_2(\mathbf{W}_2[\sigma_1(\mathbf{W}_1 \mathbf{X}_{sS}^{[l]})]) \tag{2}$$

Where $\mathbf{F}_{cE}(\cdot)$ is the excitation operator, $\mathbf{X}_{cE,i}^{[l]}$ is the intermediate excitation, $\sigma_1(\cdot)$ and $\sigma_2(\cdot)$ are rectified linear unit (ReLU) [31] and sigmoid function, respectively, with $\mathbf{W}_1 \in \mathbb{R}^{\frac{C}{r} \times C}$ and $\mathbf{W}_2 \in \mathbb{R}^{C \times \frac{C}{r}}$. The intermediate excitation performs dimension reduction specified by a reduction ratio, $r$ that determines the capacity and computational cost of the cSE block. The dimension is increased back and maps to between 0 and 1 using a sigmoid function. This is then used to recalibrate $\mathbf{X}_i^{[l]}$ by doing channel-wise multiplication and can be expressed as $\mathbf{X}_{cSE}^{[l]} = \mathbf{X}_i^{[l]} \cdot \mathbf{X}_{cE}^{[l]}$.

In the sSE block, the channel squeeze, $\mathbf{X}_{cS}^{[l]}$ is achieved by performing a convolution on $\mathbf{X}_i^{[l]}$ with a 1×1 kernel and one output channel, and can be expressed as $\mathbf{X}_{cS}^{[l]} = \mathbf{F}_{cv}(\mathbf{X}_i^{[l]}, \mathbf{W}_{cS})$, where $\mathbf{F}_{cv}(\cdot)$ is a convolution operator and $\mathbf{W}_{cS} \in \mathbb{R}^{1 \times 1 \times C \times 1}$. $\mathbf{X}_{cS}^{[l]} \in \mathbb{R}^{H \times W}$ is a projected output along the channel axis on which a sigmoid activation is applied to get spatially excited, $\mathbf{X}_{sE}^{[l]}$, which can be written as $\mathbf{X}_{sE}^{[l]} = \sigma_2(\mathbf{X}_{cS}^{[l]})$. This spatially excited $\mathbf{X}_{sE}^{[l]}$ is then used to recalibrate $\mathbf{X}_i^{[l]}$ to generate the *sSE* block output which can be expressed as $\mathbf{X}_{sSE}^{[l]} = \mathbf{X}_i^{[l]} \cdot \mathbf{X}_{sE}^{[l]}$. Finally, $\mathbf{X}_{scSE}^{[l]}$ is achieved by $\mathbf{X}_{scSE}^{[l]} = \mathbf{X}_{cSE}^{[l]} \odot \mathbf{X}_{sSE}^{[l]}$, where $\odot$ is an aggregation operation and can be either max-out (taking maximum for a specific location), addition, multiplication, or concatenation.

The max-out operation provides competitiveness between these two SE blocks by outputting the maximum for a given location (x, y, c). So, the final excitation is formed by selectively collecting the spatial and channel excitation. The addition operation adds these two blocks elementwise. One important aspect of the addition operation is that instead of ignoring one block it provides equal importance to both of them. The multiplication operation multiplies the SE blocks elementwise. The concatenation operation concatenates them along the channel axis. In this paper, we mainly focus on max-out and additive. Because in multiplication, the final excited pixels will be those that were excited by both SE blocks. If one SE block's excitation is close to 0 and the other one's close to 1, then the resultant excitation will be near 0. So, there is a good chance of information being lost which could be crucial for tasks like segmenting wound regions. On the other hand, though in concatenation, all information is preserved, it doubles the number of channels in the final output, consequently increasing the model complexity as the subsequent convolutional layers need to process feature maps with more channels. So, considering all these facts, to utilize the benefits of max-out and addition, we aggregate both by creating parallel branches of two *scSE* modules – one aggregated by max-out and another by addition. The final parallel *scSE* (*P-scSE*) module is formed by adding these two branches elementwise. We did not do further recalibration as it has already been done twice. So, the output of P-scSE can be written as $\mathbf{X}_{p-scSE}^{[l]} = \mathbf{X}_{scSE-maxout}^{[l]} \oplus \mathbf{X}_{scSE-addition}^{[l]}$, where $\oplus$ is an elementwise addition. Figure 2 demonstrates the parallel scSE. The final output of the *l*-th decoder level is:

$$\mathbf{X}^{[l]} = \sigma_1\left(\mathbf{F}_{BN}\left[\mathbf{F}_{cv}(\mathbf{X}_{p-scSE}^{[l]}, W_{p-scSE})\right]\right) \tag{3}$$

Where $\mathbf{F}_{cv}(\cdot)$, $\mathbf{F}_{BN}(\cdot)$, and $\sigma_1(\cdot)$ are the convolutional layer, batch normalization layer, and ReLU activation function, respectively. A switch (SW) is used to form the shorted P-scSE by bypassing the max-out scSE and is used when there is a smaller number of feature maps.

**Table 1** List of augmentations. (B&C means brightness and contrast)

| Augmentation (Probability, p=0.9) | | | |
|---|---|---|---|
| Set 1 (p=0.5) | Set 2 (p=0.9) | Set 3 (p=0.2) | Set 4 (p=0.2) |
| H. Flip (p=0.8) | Scale (limit=0.5, p=1) | Perspective (p=1) | Clahe (p=1) |
| V. Flip (p=0.4) | Rotate (limit=30, p=1) | Gaussian noise (p=1) | B&C (limit=0.2, p=1) |
|  | Shift (limit=0.1, p=1) | Sharpen (p=1) | Gamma (p=1) |
|  | Combine all (p=1) | Blur (limit=3, p=1) | Hue saturation (p=1) |
|  |  | M. blur (limit=3, p=1) |  |

### 3.2.3 Loss function

We use a hybrid loss function consisting of dice loss and focal loss both having equal weights. Cross-entropy loss has the drawback of discretely computing per-pixel loss without taking into account whether or not the surrounding pixels are ground truth pixels, thereby ignoring the global scenario. Dice loss, originating from the Sørensen–Dice coefficient, on the other hand, considers information loss both locally and globally. Dice loss can be expressed as $DL = (1 - DSC)$, where $DSC$ is the dice coefficient. Focal loss (FL) comes in handy when there is a class imbalance (for instance, background >> foreground) [32]. It down-weights easy examples and focuses training on hard (misclassified) examples or false negatives using a modulating factor, $(1 - p_t)^\gamma$, and can be expressed as:

$$FL(p_t) = -\alpha_t(1-p_t)^\gamma \log(p_t) \qquad (4)$$

Where, $\gamma > 1$ is the focusing parameter, and $\alpha_t \in [0,1]$ is a weighting factor. So, the final loss function is, $L = DL + FL$.

### 3.2.4 x-FUSegNet

For the FUSeg challenge 2021 where images contain wound regions in complex backgrounds, we adopt ensembling through k-fold cross-validation on the FUSegNet. The resulting ensemble network is termed x-FUSegNet (pronounced 'cross FUSegNet'). Note that, the chronic wound dataset contains cropped images containing mainly the wound region removing the complex background. So, for the FUSeg challenge, the dataset is first split into 5 folds and then trained with the FUSegNet model 5 times keeping one fold out for validation. Thus, we obtain 5 trained models which are ensembled during the inference. The ensembled output is the average of predictions achieved from these 5 models. Such an ensemble boosts the segmentation performance. The final binary output is generated by thresholding predictions to 1 if it is greater than or equal to 0.5, otherwise 0.

### 3.2.5 Training and inference

All experiments are done on a 64-bit Ubuntu PC with an 8-core 3.4 GHz CPU and a single NVIDIA RTX 2080Ti GPU. Weight update is done using Adam optimizer [33] with an initial learning rate of $1\times10^{-4}$ and weight decay of $1\times10^{-5}$ to reduce losses. The ReduceLROnPlateau, a learning rate scheduling technique, is used to decrease the learning rate when the specified metric stops improving for a period longer than the permitted patience number. We set 0.1 and 10 as the decreasing factor and patience, respectively. Images are standardized, while ground truths are normalized first. Then a set of augmentations is performed before feeding to the network. Table 1 lists all the augmentations performed with their probabilities of being selected. With a batch size of 2, we train each model for 200 epochs while monitoring the validation loss and intersection-over-union (IoU) score. We keep storing and overwriting the checkpoint whenever the validation loss decreases or the IoU score increases. Therefore, only the best checkpoint is evaluated during inference. To avoid needless training, an early stopping with patience 30 is utilized.

### 3.2.6 Evaluation metrics

The performance of an image segmentation model is frequently assessed in the medical image segmentation community using the dice coefficient (DSC). Additionally, we assess our model using precision, recall, and intersection-over-union (IoU). We evaluate both data-based and image-based metrics. We calculate Pratt's figure of merits, PFOM [34] to evaluate the boundary performance. Here are each definition's details:

$$Precision_{data} = \frac{\sum_{i=1}^{N} TP_i}{\sum_{i=1}^{N} TP_i + \sum_{i=1}^{N} FP_i} \tag{5}$$

$$Recall_{data} = \frac{\sum_{i=1}^{N} TP_i}{\sum_{i=1}^{N} TP_i + \sum_{i=1}^{N} FN_i}s \tag{6}$$

$$DSC_{data} = \frac{\sum_{i=1}^{N} 2TP_i}{\sum_{i=1}^{N} 2TP_i + \sum_{i=1}^{N} FP_i + \sum_{i=1}^{N} FN_i} \tag{7}$$

$$IoU_{data} = \frac{\sum_{i=1}^{N} TP_i}{\sum_{i=1}^{N} TP_i + \sum_{i=1}^{N} FP_i + \sum_{i=1}^{N} FN_i} \tag{8}$$

$$Precision_{image} = \frac{1}{N} \sum_{i=1}^{N} \frac{TP_i}{TP_i + FP_i} \tag{9}$$

$$Recall_{image} = \frac{1}{N} \sum_{i=1}^{N} \frac{TP_i}{TP_i + FN_i} \tag{10}$$

$$DSC_{image} = \frac{1}{N} \sum_{i=1}^{N} \frac{2TP_i}{2TP_i + FP_i + FN_i} \tag{11}$$

$$IoU_{image} = \frac{1}{N} \sum_{i=1}^{N} \frac{TP_i}{TP_i + FP_i + FN_i} \tag{12}$$

$$PFOM = \frac{1}{\max(I_{gb}, I_{pb})} \sum_{i=1}^{I_{pb}} \frac{1}{1 + \beta d(i)^2} \tag{13}$$

Here *TP*, *FP*, *TN*, and *FN* are true positive, false positive, true negative, and false negative, respectively. The number of boundary points in the ground truth and prediction, respectively, are $I_{gb}$ and $I_{pb}$. In order to establish a relative penalty between smeared and isolated borders, a scaling constant, $\beta$ selected to be 1/9 as reported in [34]. $d(i)$ is the pixel miss distance of the *i*th edge detected. In other words, it is the pixel Euclidean distance of the *i*th boundary point between the ground truth and the prediction.

Table 2 Segmentation result on the Chronic Wound dataset. Results in the first section are taken from [15] and [19]. 2nd and 3rd sections tabulated the results that we performed on state-of-the-art methods. 4th section lists the results obtained by our proposed model.

| Model | Image-based | | | | Data-based | | | | Param (M) |
|---|---|---|---|---|---|---|---|---|---|
| | IoU | P | R | DSC | IoU | P | R | DSC | |
| VGG16 | NA | NA | NA | NA | NA | 83.91 | 78.35 | 81.03 | 134.3 |
| SegNet | NA | NA | NA | NA | NA | 83.66 | 86.49 | 85.05 | **0.90** |
| Mask RCNN | NA | NA | NA | NA | NA | 94.30 | 86.40 | 90.20 | 63.62 |
| MobileNetV2+CCL | NA | NA | NA | NA | NA | 91.01 | 89.97 | 90.47 | 2.14 |
| LinkNet-EffB1 + UNet-EffB2 | NA | NA | NA | 84.42 | 85.51 | 92.68 | **91.80** | 92.07 | NA |
| MANet [35] | 76.97 | 86.28 | 83.85 | 83.80 | 84.71 | 93.11 | 90.37 | 91.72 | 76.35 |
| FPN [36] | 76.40 | 86.77 | 82.94 | 83.37 | 83.64 | 92.52 | 89.71 | 91.09 | 65.67 |
| TransUNet [37] | 75.61 | 85.38 | 82.50 | 82.61 | 81.15 | 92.07 | 87.25 | 89.59 | 16.97 |
| LinkNet [38] | 74.36 | 87.12 | 80.71 | 81.89 | 82.74 | 94.24 | 87.15 | 90.56 | 62.78 |
| PSPNet [39] | 75.09 | 85.40 | 82.84 | 82.65 | 84.63 | 92.57 | 90.79 | 91.67 | 1.02 |
| DeepLabV3Plus [40] | 77.02 | 86.04 | 84.91 | 83.96 | 85.19 | 92.75 | 91.27 | 92.00 | 63.46 |
| Swin-Unet [41] | 63.57 | 76.83 | 74.74 | 73.15 | 79.30 | 89.94 | 87.02 | 88.46 | 57.45 |
| MiT-b5-Unet [42], [43] | 77.48 | 87.83 | 83.66 | 84.28 | 85.27 | 93.79 | 90.38 | 92.05 | 84.72 |
| DDRNet [44] | 53.12 | 74.88 | 62.85 | 63.73 | 57.64 | 80.86 | 66.75 | 73.13 | 5.7 |
| SegFormer-b5 [42] | 71.82 | 84.67 | 78.42 | 79.57 | 83.58 | 92.21 | 89.94 | 91.06 | 81.97 |
| U-Net | 77.92 | 86.58 | 85.31 | 84.66 | 83.28 | 90.31 | 91.45 | 90.88 | 65.45 |
| U-Net + scSE (additive) | 77.81 | **88.72** | 83.35 | 84.15 | 84.92 | **94.89** | 89.00 | 91.85 | 65.46 |
| U-Net + scSE (max-out) | 78.21 | 88.06 | 84.11 | 84.75 | 85.15 | 94.62 | 89.48 | 91.98 | 65.46 |
| Proposed FUSegNet | **79.44** | 88.29 | **86.35** | **86.05** | **86.40** | 94.40 | 91.07 | **92.70** | 64.90 |

*P and R mean precision and recall, respectively

Table 3 Top five performers of the MICCAI 2021 FUSeg Challenge [27].

| Team | Approach | Data-based DSC (%) |
|---|---|---|
| Mrinal et al. | x-FUSegNet | **89.23** |
| Mahbod et al. | LinkNet-EffB1 + UNet-EffB2 | 88.80 |
| Zhang et al. | U-Net with HarDNet68 | 87.57 |
| Galdran et al. | Stacked U-Nets | 86.91 |
| Hong et al. | NA | 86.27 |

**Table 4** Average test time in seconds taken by models achieving more than 90% DSC in the 2nd section of Table 2 during inference.

|  | MANet | LinkNet | DeepLabV3+ | MiT-b5-Unet | SegFormer | FUSegNet | x-FUSegNet |
|---|---|---|---|---|---|---|---|
| Exe. Time (s) | 0.034 | 0.032 | 0.032 | 0.040 | 0.043 | 0.033 | 0.175 |

## 4 Results and Discussion

### 4.1 Results

Models are primarily evaluated by intersection-over-union (IoU), precision, recall, and dice coefficient (DSC). First, an extensive study is performed on the chronic wound dataset as its test images are publicly available with corresponding masks. Results for state-of-the-art methods are listed in Table 2. The table is split into four sections. The first section lists the results for the same dataset reported in the published research papers. In the second and third sections in Table 2, we generate outputs using different state-of-the-art methods. The third section, particularly, lists the performance of different scSE modules fused in U-Net architecture. The last section tabulates the performance of our proposed FUSegNet model. Our proposed model outperforms the existing approaches and achieves a DSC of 86.05 for image-based and 92.70 for data-based evaluation. Figure 3 demonstrates some outputs generated by the FUSegNet for different DSC scores. Figure 4 demonstrates how the P-scSE module is working in the decoding process. For further analysis, images are categorized into 10 groups based on the size of the ulcer. Figure 5 shows the boxplot representation of the image-based evaluation of each category. In addition, we use a pie chart to demonstrate the data-based evaluation of each category. For large ulcer regions, we achieve DSC scores of more than 90%. Figure 6 plots the DSC scores and PFOM scores for SE-fused models for different categories. It is seen that the P-scSE-fused FUSegNet outperforms the other models in most of the categories. Table 3 shows the top five teams currently on the MICCAI FUSeg Challenge 2021 leaderboard with our x-FUSegNet being at the top of it. Table 4 lists the average time taken by models to make predictions during inference. Model loading time is not included here, as it is loaded only once during the inference process. Figure 7 and Figure 8 illustrate the prediction result for the x-FUSegNet in the FUSeg Challenge dataset. We then analyze the predictions from visual inspection as the test data is not publicly available.

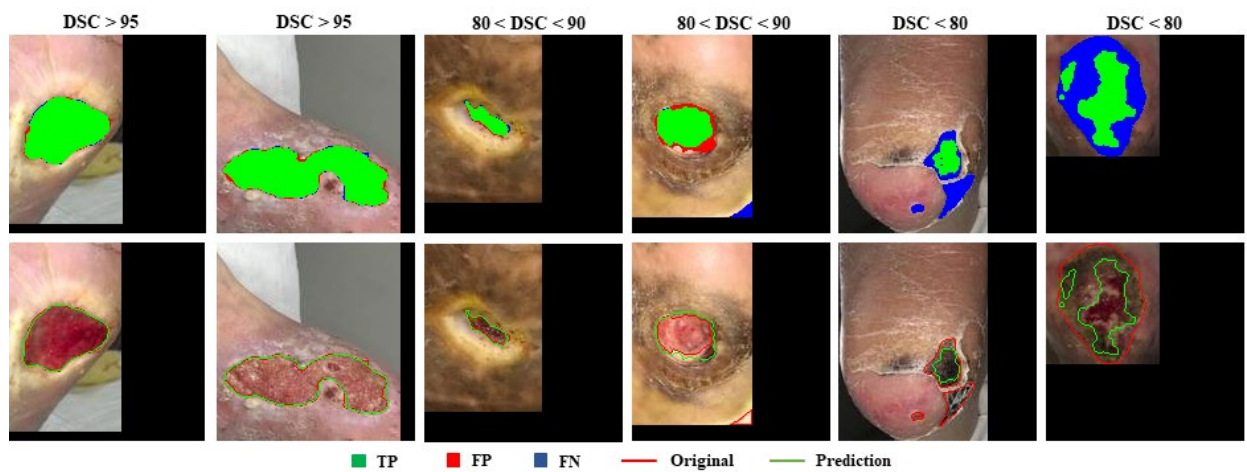

**Figure 3** Segmentation results achieved by the FUSegNet for Chronic Wound dataset for different DSC scores. (top) TP, FP, and FN regions are marked with green, red, and blue, respectively. (bottom) The original and predicted boundaries of the ulcer regions are shown in red and green colors, respectively.

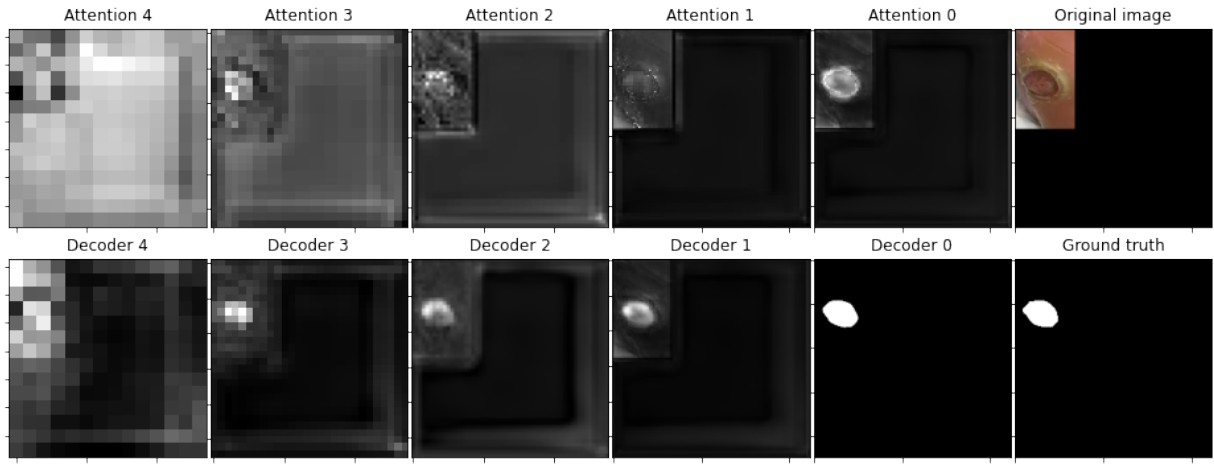

**Figure 4** Demonstration of how the prediction formed in the decoding process. (*top*) Outputs right after applying the P-scSE module. (*bottom*) Outputs of each decoder stage. Outputs are taken applying conv2D-ReLU-BN after performing P-scSE except for the top layer (Decoder 0) where additional conv2d and sigmoid activation function is applied to create the final output.

### 4.2 Discussion

The main objective of this work is to develop a deep-learning model for the FUSeg Challenge 2021 that contains diabetic foot ulcer images. Since the segmentation masks for the test images of the chronic wound dataset are publicly available but those for the test images of the FUSeg dataset are not, we first conduct a thorough analysis of the chronic wound dataset, a publicly available DFU dataset, to develop the network. First, we build an encoder-decoder-based model where a pretrained EfficientNet-b7 is used as the backbone. EfficientNet seeks to achieve a balanced depth, width, and scale resolution which is very important in medical image segmentation. We propose a new decoder architecture by using a modified spatial and channel squeeze-and-excitation module that we call P-scSE. The name comes from the fact that two parallel branches of additive and max-out scSE are combined together. In the final decoder stage, a modified P-scSE called shorted P-scSE is used. The intuition is that the final decoder stage has a very limited number of feature maps to process. So, taking max-out will risk losing important features. Instead, we bypass the max-out scSE and feed the feature maps from the decoder input at the P-scSE module's output, where they are combined with the additive scSE output.

From Table 2, it is seen that our proposed model outperforms the state-of-the-art methods for both image-based and data-based evaluation. Figure 3 shows the segmentation results for different DSC scores. We also explore the following transformer-based models: Swin-Unet, Mixed Vision Transformer (MiT)-b5-Unet, and SegFormer-b5. Among them, MiT-b5-Unet showed better performance with a data-base DSC of 92.05%, albeit with an expense of 84.72 million parameters. In MiT-b5-Unet, we use the encoder from SegFormer-b5 and the decoder from the U-Net architecture. In addition, we compare our proposed model with the U-Net architecture fused with the scSE module. Roy et al. [26] proposed the scSE module at the end of each decoder stage in U-Net architecture. In contrast, we fuse our P-scSE module in the middle followed by a Conv-BN-ReLU block. Figure 4 shows that such an arrangement smooths the attention output and sends a well-represented feature map to the next decoder stage. It also outperforms the scSE-based score. It is also notable that our proposed model has a reduced number of parameters than the scSE-fused U-Net and has a very competitive number of parameters when compared to most of the state-of-the-art methods.

For further analysis, as shown in Figure 5(b), we divide the test images into 10 different categories based on the no. of ground truth (GT) pixels. For instance, "%GT area < 0.15" means that the no. of foot ulcer pixels is less than 0.15% compared to the total no. of pixels in that particular image. Similarly, "%GT area 0" means there is no foot ulcer in the image. We then perform an image-based evaluation on each category to generate boxplots. Figure 5(a) shows the boxplot representation of each category. The whiskers endpoints are selected as the (1st quartile - 1.5 × interquartile range (IQR)) and (3rd quartile + 1.5 × IQR). A green triangle symbolizes the mean, while an orange line represents the median or the 2$^{nd}$ quartile ($Q_2$). The evaluation metrics are displayed on the x-axis label. The red dots are the outliers. Outliers are the points that are outside the range of the whisker interval. It is seen that in most cases, $Q_2$ lies at the upper half of the interquartile box. The $Q_2$ and mean value indications, which approach 90% or more as the foot ulcer region increases, show that overall performances for assessment measures are often over 80% even when certain outliers, particularly for categories 2 and 4, fall to very low values. It is very difficult to interpret the boxplot for Category 1 since it does not have any foot ulcer region. So, even a very tiny region in the prediction will result in zero DSC. Hence, the DSC will either be 0 or 100% for category 1. So, as shown in Figure 5(d), another boxplot is generated that represents the no. of non-zero intensities found in category 1. Images in this category do not have any foot ulcers. So, any presence of non-zero intensity in the prediction is a false-positive (FP). The mean and median are close to 100 and 50, respectively, which is quite low compared to the total no. of pixels in an image.

We also draw a pie chart as shown in Figure 5(c) that indicates the DSC for each category. This time a data-based evaluation is performed. We first calculate TP, FP, and FN for a specific category, and then calculate the DSC of that particular category. We exclude Category 1 as it does not contain any foot ulcer pixels. It can be inferred that the DSC score improves as the area with foot ulcers grows. For categories 6-10, DSC is above 90%, while for categories 3-5, it is equal to or close to 90%. Additionally, we assess how well our model performs in comparison to the scSE-fused U-Net models in terms of DSC scores. Figure 6(left) demonstrates the proposed model outperforms the other models in most of the categories.

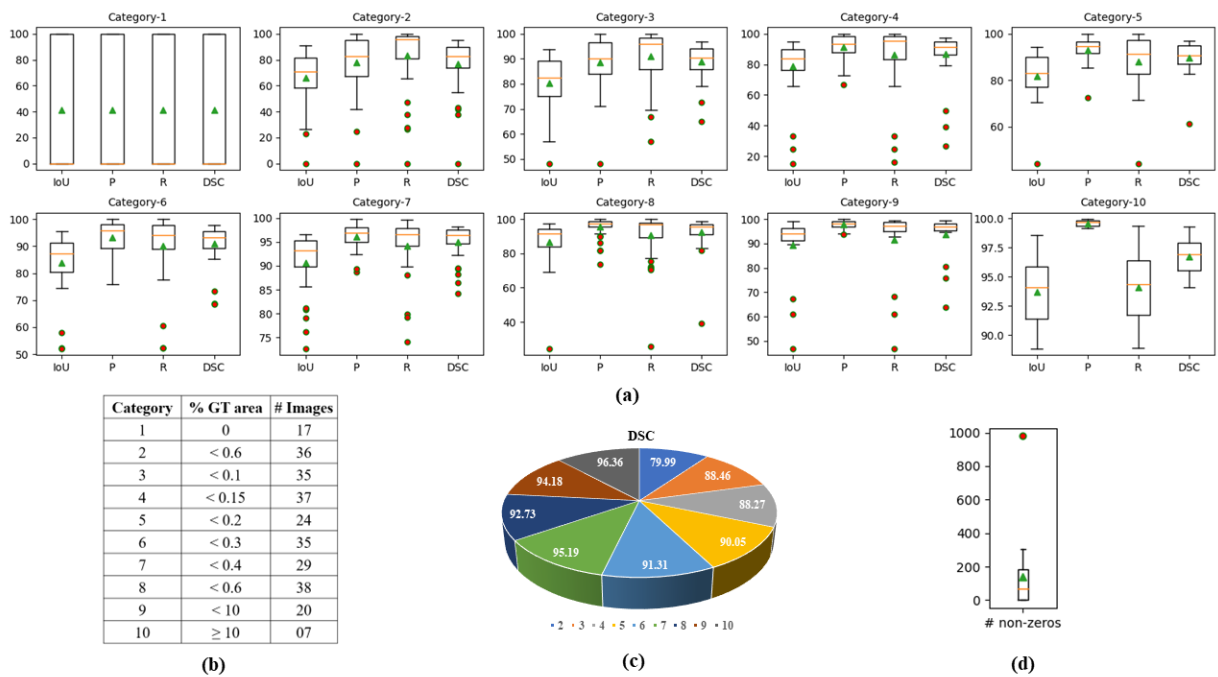

**Figure 5** (*a*) Image-based evaluations for each category for the Chronic Wound dataset are demonstrated using boxplots. (*b*) Specification of category. Categories are generated based on the percentage ground truth area. (*c*) Pie-chart representation of data-based evaluation for each category. (*d*) Boxplot representation of no. of non-zero intensity found in category 1. Images in this category do not have any foot ulcers. So, any presence of non-zero intensity in the prediction is a false-positive (FP).

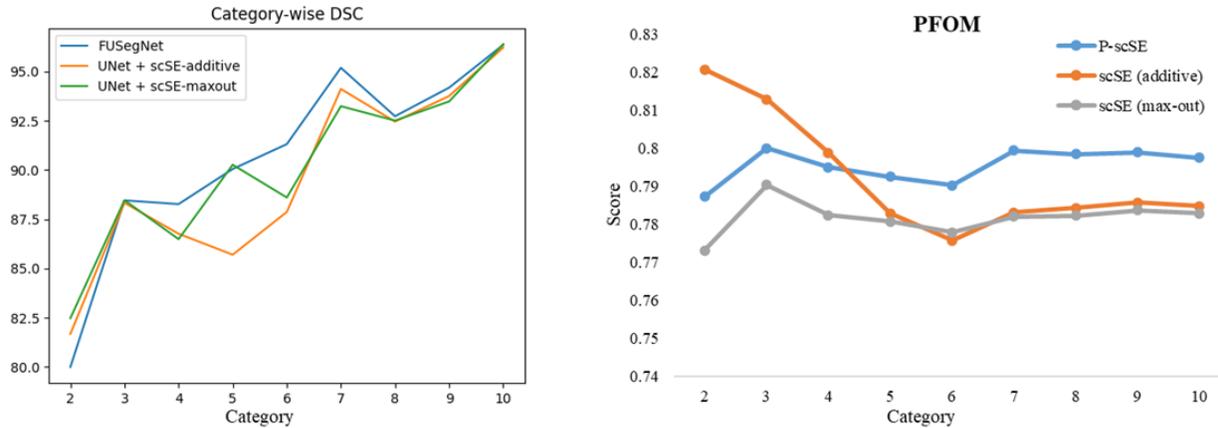

**Figure 6** (*left*) DSC scores for the Chronic Wound dataset using different squeeze-and-excitation (SE) modules for different categories. (*right*) PFOM scores to evaluate the boundary performance for different categories.

We then perform a contour test. The goal is to measure how close the ground truth and prediction contours are. To do that, we use Pratt's figure of merits (PFOM). The PFOM index evaluates the accuracy of edge localization. We use Canny edge detection with a Gaussian kernel that has an extremely low standard deviation (0.1) to prevent over-smoothing. Also, when there is no edge in any of the images, then PFOM will be infinite. To avoid computational complexity, we replace infinity with a relatively high value. In our case, we set it to 2. We then compare the PFOM scores for different squeeze-and-excitation (SE) modules. Figure 6(right) shows that the scSE (additive) works well when the ulcer area is very small, but it falls rapidly as the region grows. On the other hand, our proposed FUSegNet shows consistent performance while dominating most of the categories. It is found that the scSE (max-out) performs the least well of these three mechanisms.

After performing extensive tests on the chronic wound dataset to finalize all the parameters for the FUSegNet model, we then apply it to the FUSeg Challenge 2021. The only change made is the submission of the x-FUSegNet, a variant of FUSegNet, which trains the FUSegNet model five times using 5-fold cross-validation. The final output is taken by doing a pixel-wise average of all 5 models. We ensemble these models so that a better performance can be achieved in complex backgrounds. Table 3 lists the results for the top 5 approaches in the MICCAI FUSeg Challenge 2021. The organizer evaluates the data-based metrics only and ranks based on the dice score. Currently, our model is at the top of the leaderboard [27].

As the test image masks of the FUSeg Challenge are not publicly available, we manually analyze the model's output. Figure 7 shows some outputs that we think are quite good. The segmented regions are marked with green color. Visual inspection suggests that the model separates the foot ulcer area from the skin around the foot. In certain images, we also discover some anomalies. Figure 8 illustrates a few such instances. In Figure 8(a-b), after visual inspection, we find that the model has detected some regions that are actually not foot ulcer regions. Such areas have a color intensity that is roughly a blend of the wound and nearby skin color. It only identifies one of the two wound regions in Figure 8(c). In Figure 8(d-e), the model misses some portions of the wound regions. In these two cases, the inner portion of the wound region has a relatively deeper intensity and gradually fades away as it approaches the contour of the wound. The model produces a false negative considering these faded areas as the surrounding foot skin. Deep visual investigation reveals that the model appears to function very well for the extensive callus and deep ulcer areas including granulation, necrotic, and eschar tissues, however, there is room for improvement in the superficial ulcer.

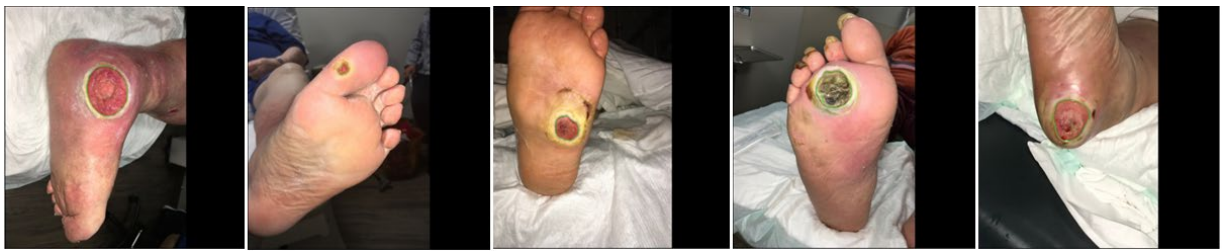

**Figure 7** Segmentation results for the FUSeg dataset that appear promising after performing manual inspections. Boundaries of the predicted region are marked with green color.

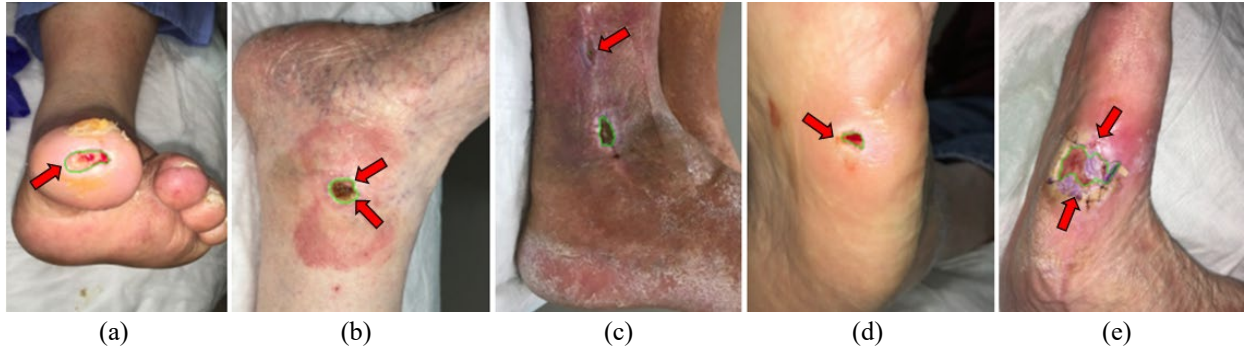

(a)            (b)            (c)            (d)            (e)

**Figure 8** Segmentation results for the FUSeg dataset that appear problematic after performing manual inspections. Arrows are showing the spots where problems are found.

In Table 4, we present the average time taken by models to generate predictions. We consider models that achieved more than a 90% data-based DSC as listed in the second section of Table 2. The execution time taken by FUSegNet is almost the same as MANet, LinkNet, and DeepLabV3+, with around 0.03 seconds, and lower than transformer-based models which take around 0.04 seconds. Even our ensemble-based model (x-FUSegNet) takes less than 0.2 seconds. Therefore, considering the number of parameters (64.90M) and the execution time (0.033 seconds), our proposed FUSegNet achieves an advanced balance between computational complexity and segmentation performance compared to other state-of-the-art models.

## 5    Conclusion

Untreated chronic wounds are costly and negatively impact patients' quality of life. Diabetes and obesity increase the risk of chronic wounds, such as foot ulcers, which can lead to serious consequences. Proper identification and monitoring of the wound area are crucial for effective treatment, but manual extraction is time-consuming and expensive. A fully automated computer-aided foot ulcer segmentation method is a faster and more effective solution.

In this paper, we propose a deep learning-based fully automatic encoder-decoder model called FUSegNet to segment the foot ulcer region. We construct a modified spatial and channel squeeze-and-excitation module called P-scSE that we fuse to the middle of each decoder stage. The model is primarily developed for the FUSeg Challenge 2021. Before applying it, we perform several experiments on the chronic wound dataset. Finally, we come up with x-FUSegNet that ensembles the outputs of the FUSegNet trained using 5-fold cross-validation. Our proposed model outperforms the current approaches for the FUSeg Challenge 2021 in terms of the dice score.

In this paper, only the ulcer region is considered. The ulcer region can be divided into different sub-regions based on the presence of different tissues. So, future works include zonal segmentation by characterizing different tissues present in the ulcer region. Precise evaluation and monitoring of various tissue types in chronic wounds are essential for managing and tracking the wound's healing progress. This information helps to plan future treatments for the wound.

**REFERENCE**

[1] S. Bowers and E. Franco, "Chronic wounds: Evaluation and management," *American Family Physician*, vol. 101, no. 3, pp. 159–166, 2020.

[2] M. Chang and T. T. Nguyen, "Strategy for Treatment of Infected Diabetic Foot Ulcers," *Accounts of Chemical Research*, vol. 54, no. 5, pp. 1080–1093, 2021.

[3] A. J. M. Boulton, "The diabetic foot," *Medicine (United Kingdom)*, vol. 47, no. 2, pp. 100–105, 2019.

[4] G. Scebba et al., "Detect-and-segment: A deep learning approach to automate wound image segmentation," *Informatics in Medicine Unlocked*, vol. 29, no. February, pp. 100884, 2022.

[5] S. R. Nussbaum et al., "An Economic Evaluation of the Impact, Cost, and Medicare Policy Implications of Chronic Nonhealing Wounds," *Value in Health*, vol. 21, no. 1, pp. 27–32, 2018.

[6] L. B. Jørgensen, J. A. Sørensen, G. B. E. Jemec, and K. B. Yderstræde, "Methods to assess area and volume of wounds – a systematic review," *International Wound Journal*, vol. 13, no. 4, pp. 540–553, 2016.

[7] C. K. Sen, "Human Wound and Its Burden: Updated 2020 Compendium of Estimates," *Advances in Wound Care*, vol. 10, no. 5, pp. 281–292, 2021.

[8] B. Song and A. Sacan, "Automated wound identification system based on image segmentation and artificial neural networks," *Proceedings - 2012 IEEE International Conference on Bioinformatics and Biomedicine, BIBM 2012*, pp. 619–622, 2012.

[9] K. Wantanajittikul, S. Auephanwiriyakul, N. Theera-Umpon, and T. Koanantakool, "Automatic segmentation and degree identification in burn color images," *BMEiCON-2011 - 4th Biomedical Engineering International Conference*, pp. 169–173, 2011.

[10] M. Jawahar, L. Jani Anbarasi, S. Graceline Jasmine, and M. Narendra, "Diabetic foot ulcer segmentation using color space models," *Proceedings of the 5th International Conference on Communication and Electronics Systems, ICCES 2020*, pp. 742–747, 2020.

[11] S. Thomas, "Medetec Wound Database". [Online]. Available: http://www.medetec.co.uk/files/medetec-image-databases.html. (Last accessed: May 2023)

[12] A. Heras-Tang, D. Valdes-Santiago, A. Leon-Mecias, M. L. B. Diaz-Romañach, and J. A. Mesejo-Chiong, "Diabetic foot ulcer segmentation using logistic regression, DBSCAN clustering and morphological operators," *Electronic Letters on Computer Vision and Image Analysis*, vol. 21, no. 2, pp. 23–39, 2022.

[13] M. Goyal, N. D. Reeves, S. Rajbhandari, J. Spragg, and M. H. Yap, "Fully convolutional networks for diabetic foot ulcer segmentation," *2017 IEEE International Conference on Systems, Man, and Cybernetics, SMC 2017*, pp. 618–623, 2017.

[14] X. Liu, C. Wang, F. Li, X. Zhao, E. Zhu, and Y. Peng, "A framework of wound segmentation based on deep convolutional networks," *Proceedings - 2017 10th International Congress on Image and Signal Processing, BioMedical Engineering and Informatics, CISP-BMEI 2017*, pp. 1–7, 2018.

[15] C. Wang et al., "Fully automatic wound segmentation with deep convolutional neural networks," *Scientific Reports*, vol. 10, no. 1, 2020.

[16] C. Cao et al., "Nested segmentation and multi-level classification of diabetic foot ulcer based on mask R-CNN," *Multimedia Tools and Applications*, vol. 82, no. 12, pp. 18887-18906, 2022.

[17] D. Ramachandram, J. L. Ramirez-GarciaLuna, R. D. J. Fraser, M. A. Martínez-Jiménez, J. E. Arriaga-Caballero, and J. Allport, "Fully Automated Wound Tissue Segmentation Using Deep Learning on Mobile Devices: Cohort Study," *JMIR mHealth and uHealth*, vol. 10, no. 4, pp. 1–19, 2022.

[18] H. N. Huang et al., "Image segmentation using transfer learning and Fast R-CNN for diabetic foot wound

treatments," *Frontiers in Public Health*, vol. 10, no. 1, 2022.

[19] A. Mahbod, G. Schaefer, R. Ecker, and I. Ellinger, "Automatic Foot Ulcer Segmentation Using an Ensemble of Convolutional Neural Networks," *Proceedings - International Conference on Pattern Recognition*, vol. 2022-August, pp. 4358–4364, 2022.

[20] C. Wang *et al.*, "FUSeg: The Foot Ulcer Segmentation Challenge," pp. 1–14, 2022. [Online]. Available: http://arxiv.org/abs/2201.00414

[21] C. Kendrick *et al.*, "Translating Clinical Delineation of Diabetic Foot Ulcers into Machine Interpretable Segmentation," pp. 1–11, 2022. [Online]. Available: http://arxiv.org/abs/2204.11618

[22] H. Yi *et al.*, "OCRNet for Diabetic Foot Ulcer Segmentation Combined with Edge Loss," *Lecture Notes in Computer Science (including subseries Lecture Notes in Artificial Intelligence and Lecture Notes in Bioinformatics)*, vol. 13797 LNCS, pp. 31–39, 2023.

[23] M. Hassib, M. Ali, A. Mohamed, M. Torki, and M. Hussein, "Diabetic Foot Ulcer Segmentation Using Convolutional and Transformer-Based Models," *Lecture Notes in Computer Science (including subseries Lecture Notes in Artificial Intelligence and Lecture Notes in Bioinformatics)*, vol. 13797 LNCS, pp. 83–91, 2023.

[24] A. S. Y. Lien, C. Y. Lai, J. Da Wei, H. M. Yang, J. T. Yeh, and H. C. Tai, "A Granulation Tissue Detection Model to Track Chronic Wound Healing in DM Foot Ulcers," *Electronics (Switzerland)*, vol. 11, no. 16, 2022.

[25] A. Kairys, R. Pauliukiene, V. Raudonis, and J. Ceponis, "Towards Home-Based Diabetic Foot Ulcer Monitoring: A Systematic Review," *Sensors*, vol. 23, no. 7, pp. 1–29, 2023.

[26] A. G. Roy, N. Navab, and C. Wachinger, "Recalibrating Fully Convolutional Networks With Spatial and Channel 'Squeeze and Excitation' Blocks," *IEEE Transactions on Medical Imaging*, vol. 38, no. 2, pp. 540–549, 2019.

[27] C. Wang, "Foot Ulcer Segmentation Challenge 2021." [Online]. Available: https://fusc.grand-challenge.org (Last accessed: May, 2023)

[28] J. Redmon and A. Farhadi, "YOLOv3: An Incremental Improvement," 2018. [Online]. Available: http://arxiv.org/abs/1804.02767

[29] M. Tan and Q. V. Le, "EfficientNet: Rethinking model scaling for convolutional neural networks," *36th International Conference on Machine Learning, ICML 2019*, vol. 2019-June, pp. 10691–10700, 2019.

[30] J. Hu, "Squeeze-and-Excitation Networks," in *Proc. - IEEE conference on computer vision and pattern recognition*, 2018, pp. 7132–7141.

[31] G. E. H. Vinod Nair, "Rectified Linear Units Improve Restricted Boltzmann Machines," in *Proceedings of the 27th International Conference on Machine Learning, Haifa, Israel*, 2010.

[32] T. Y. Lin, P. Goyal, R. Girshick, K. He, and P. Dollar, "Focal Loss for Dense Object Detection," *IEEE Transactions on Pattern Analysis and Machine Intelligence*, vol. 42, no. 2, pp. 318–327, 2020.

[33] D. P. Kingma and J. L. Ba, "Adam: A method for stochastic optimization," *3rd International Conference on Learning Representations, ICLR 2015 - Conference Track Proceedings*, pp. 1–15, 2015.

[34] D. H. Lim, "Robust edge detection in noisy images," vol. 50, pp. 803–812, 2006.

[35] T. Fan, G. Wang, Y. Li, and H. Wang, "Ma-net: A multi-scale attention network for liver and tumor segmentation," *IEEE Access*, vol. 8, pp. 179656–179665, 2020.

[36] T.-Y. Lin, P. Dollár, R. Girshick, K. He, B. Hariharan, and S. Belongie, "Feature Pyramid Networks for Object Detection," *Proceedings of the IEEE Conference on Computer Vision and Pattern Recognition*, pp. 2117–2125, 2017.

[37]     J. Chen *et al.*, "TransUNet: Transformers Make Strong Encoders for Medical Image Segmentation," pp. 1–13, 2021. [Online]. Available: http://arxiv.org/abs/2102.04306

[38]     A. Chaurasia and E. Culurciello, "LinkNet: Exploiting encoder representations for efficient semantic segmentation," *2017 IEEE Visual Communications and Image Processing, VCIP 2017*, vol. 2018-Janua, pp. 1–4, 2018.

[39]     H. Zhao, J. Shi, X. Qi, X. Wang, and J. Jia, "Pyramid scene parsing network," *Proceedings - 30th IEEE Conference on Computer Vision and Pattern Recognition, CVPR 2017*, vol. 2017-Janua, pp. 6230–6239, 2017.

[40]     L. Chen, Y. Zhu, G. Papandreou, F. Schroff, and C. V Aug, "Encoder-Decoder with Atrous Separable Convolution for Semantic Image Segmentation," 2018. [Online]. Available: https://arxiv.org/abs/1802.02611

[41]     H. Cao *et al.*, "Swin-Unet: Unet-like Pure Transformer for Medical Image Segmentation," in *European conference on computer vision. Cham: Springer Nature Switzerland*, 2022, pp. 205–218.

[42]     E. Xie, W. Wang, Z. Yu, A. Anandkumar, J. M. Alvarez, and P. Luo, "SegFormer: Simple and Efficient Design for Semantic Segmentation with Transformers," *Advances in Neural Information Processing Systems*, vol. 15, no. NeurIPS, pp. 12077–12090, 2021.

[43]     T. Ronneberger, O., Fischer, P. and Brox, "U-Net: Convolutional Networks for Biomedical Image Segmentation," in *Medical Image Computing and Computer-Assisted Intervention–MICCAI 2015: 18th International Conference, Munich, Germany, October 5-9, 2015, Proceedings, Part III 18. Springer International Publishing*, 2015, pp. 234–241.

[44]     H. Pan, Y. Hong, W. Sun, and Y. Jia, "Deep Dual-Resolution Networks for Real-Time and Accurate Semantic Segmentation of Traffic Scenes," *IEEE Transactions on Intelligent Transportation Systems*, vol. 24, no. 3, pp. 3448–3460, 2023.